\documentclass[final,5p,times,twocolumn,authoryear]{elsarticle}

\usepackage{amssymb}

\usepackage{times}
\usepackage{epsfig}
\usepackage{graphicx}
\usepackage{amsmath}
\usepackage{amssymb}
\usepackage{mathtools}
\usepackage{standalone}
\usepackage{csquotes}
\usepackage[usenames, dvipsnames]{color}
\usepackage{color,soul}
\usepackage{tabularx}
\usepackage{amsmath}
\usepackage{nccmath}
\usepackage{color}
\usepackage[table,xcdraw]{xcolor}
\usepackage{float}
\usepackage{pdflscape}
\usepackage{rotating}
\usepackage{url}

\newcommand{\ie}{\textit{i}.\textit{e}.}
\newcommand{\eg}{\textit{e}.\textit{g}.}

\journal{Science of Remote Sensing}

\begin{document}

\begin{frontmatter}



\title{A Nearest Neighbor Network to Extract Digital Terrain Models from 3D Point Clouds}


\author[label1]{Mohammed Yousefhussien}
\author[label2]{David J. Kelbe}
\author[label1]{Carl Salvaggio}
\address[label1]{Rochester Institute of Technology, Chester F. Carlson Center for Imaging Science, Rochester, NY, USA}
\address[label2]{Xerra Earth Observation Institute, Alexandra, New Zealand}

\begin{abstract}
 When 3D-point clouds from overhead sensors are used as input to remote sensing data exploitation pipelines, a large amount of effort is devoted to data preparation. Among the multiple stages of the preprocessing chain, estimating the Digital Terrain Model (DTM) model is considered to be of a high importance; however, this remains a challenge, especially for raw point clouds derived from optical imagery. Current algorithms estimate the ground points using either a set of geometrical rules that require tuning multiple parameters and human interaction, or cast the problem as a binary classification machine learning task where ground and non-ground classes are found. In contrast, here we present an algorithm that directly operates on 3D-point clouds and estimate the underlying DTM for the scene using an end-to-end approach without the need to classify points into ground and non-ground cover types. Our model learns neighborhood information and seamlessly integrates this with point-wise and block-wise global features. We validate our model using the ISPRS 3D Semantic Labeling Contest LiDAR data, as well as three  scenes generated using dense stereo matching, representative of high-rise buildings, lower urban structures, and a dense old-city residential area. We compare our findings with two widely used software packages for DTM extraction, namely ENVI and LAStools. Our preliminary results show that the proposed method is able to achieve an overall Mean Absolute Error of 11.5\% compared to 29\% and 16\% for ENVI and LAStools.

\end{abstract}


\end{frontmatter}


\section{Introduction}
\label{sec:introduction}
The last decade has seen dramatic improvements in the capacity to collect and interpret 3D data from Earth observing (EO) platforms. First, advances in hardware such as Light Detection and Ranging (LiDAR) and Synthetic Aperture Radar (SAR) have enabled rapid measurement of 3D coordinates directly. And second, advances in computing have revived historical photogrammetry techniques, giving rise to scalable software packages such as dense matching and Structure from Motion (SfM), which generate 3D data indirectly from stereo or multiple-view optical imagery. This dramatic increase in the availability of 3D Earth observation data has advanced a variety of applications such as: city planning \citep{city_planning_2019}, agricultural surveying~\citep{LiDAR_Agri}, post-disaster recovery~\citep{yang2018comprehensive}, and flood flood mapping and assessment~\citep{Kabite_flood}, to name a few.

The success of these applications hinges upon complex processing in order to translate raw 3D point data into actionable information products. Typical preprocessing pipelines for point cloud data involve a series of stages such as noise filtering, feature extraction, and terrain `normalization'. Normalization involves first estimating the bare Earth terrain; these values are then subtracted from the z coordinates of the point cloud to obtain height-above-ground. This is critical in order to ensure the accuracy of subsequent applications that rely on classifying points or measuring the height of features above ground, as such algorithms are susceptible to false predictions introduced by the bias of the varying Earth surface.~\citep{my_isprs_paper,chen2018semantic}. Not to mention, there are numerous applications that require the bare-Earth terrain directly. 

Unfortunately modeling the Earth surface is a challenging task for overhead sensors, which often can only observe the uppermost surface. This is especially true for point clouds derived using optical imaging. As a result, while the Digital Surface Model (DSM)--that is, the Earth surface and all the objects that are on top of it--is simple and straightforward to generate, the so-called Digital Terrain Model (DTM) is not. The DTM includes only the bare-Earth surface and thus must exclude buildings, trees, cars, and other objects. The residual after subtracting the DTM from the DSM is known as the normalized Digital Surface Model (nDSM) or the Digital Height Model (DHM). Note that these terms most commonly refer to raster surfaces; however the same concepts apply to point data. 

Many of the point cloud applications listed above require that the DTM first be estimated so that they can operate on the normalized data. With an eye towards supporting these downstream applications, this paper presents a new method to estimate the DTM directly.  


We introduce a flexible and simple deep learning framework that directly estimates the DTM from 3D-point clouds obtained using either LiDAR or optical imagery. We extend PointNet~\citep{pointnet} to learn neighborhood features while preserving the simple and elegant network architecture. For each point, the algorithm estimates the bare Earth $z$-value directly, without the need to pre-classify the data into ground and non-ground cover types. This preserves the sparse form of the point cloud data and avoids the additional error arising from misclassification. 

\section{Related Work}
\label{sec:related_work}
In the past decades, researchers have presented a variety of methods to generate the DTM, generally taking one of the following approaches: slope-based analysis, triangulated irregular network (TIN)-based analysis, morphological filtering of a raster DSM, or more recently, machine learning. Given the large body of research in this field, here we summarize a selection of prior contributions. We refer the reader to \citet{DTM_review_Meng_2010,Chen2017StateoftheArtDG,Zhang_2010_DTM,Sithole_2003} for comprehensive reviews.

In \citet{Vosselman2k}, the author presented an algorithm to classify ground points by analyzing the slope and distance relationships between a pair of nearby points. Given a point pair, the algorithm initially hypothesizes that a large height difference between them is unlikely to be caused by a steep slope in the terrain, and asserts that the probability that the higher point is ground should increase as the distance between the points increases. Using this hypothesis, a set of geometric rules and thresholds were designed to classify all points into ground and non-ground classes. 

In the same year, \citet{axelsson2000generation} introduced a method based on an adaptive TIN model. Statistics such as minimum $z$-values within a cell are collected to seed an initial coarse TIN. The TIN model is then iteratively densified by including more points that satisfy a set of thresholds until no further points meet these criteria. The remaining points are considered non-ground points, while the points included within the TIN model are classified as ground. This method has become the base model for many commercial software packages such as TerraScan\footnote{\url{http://www.terrasolid.com/home.php}}. 

Rather than operating directly on the point cloud, other authors have opted to use a raster DSM. \citet{Wack2002} presented a hierarchical multi-scale approach for the detection of non-terrain `pixels' in a DSM. First a raster image with a large cell element is created. In this coarse representation, a Laplacian-of-Gaussian (LoG) filter is used as a blob detector to remove large non-ground objects. Next, a finer raster image is created with previously-detected non-ground objects removed. The process is repeated, iteratively, until a desired resolution; the resulting raster is the DTM. 

A similar raster-based analysis was presented by \citet{Ruijin2005}. Based on the hypothesis that smooth ground (e.g., pavement) is present in urban and suburban areas, the author approximated local regions within a window by a planar surface. This eliminated all irregular points (e.g., trees, shrubs) not belonging to ground or rooftops. These planar surfaces were then analyzed with a connected component analysis, after transforming the point cloud into a 1m grid. Using an area threshold, smaller connected regions were considered non-ground points while larger components were classified as ground points and used to interpolate a DTM. While this method may work in flat areas, the planar approximation doesn't scale well for irregular terrain.

This idea was extended by \citet{Susaki2012AdaptiveSF}, who used planar surface features and connectivity, along with local `low points' to improve DTM extraction. Point cloud data are first rasterized, and then planar surfaces such as ground and rooftops are detected. Local surface normals are analyzed to remove pitched rooftops and other features, based on the assumption that ground surfaces fall within a certain slope threshold. Several heuristics are then applied, which add ground points to the DTM model if local distance and slope relationships meet predefined thresholds. 
This slope-based paradigm was extended for photogrammetric point clouds by \citet{Misganu2016}, although they required a low-resolution DTM to seed the ground point detection.


In order to avoid heuristics based on slope or point-wise distances, another class of algorithms has proposed the use of morphological operations. This approach casts the problem as a image processing task, operating on a grayscale DSM `image' representing the height information. 

For example, \citet{MWTH2014} proposed an algorithm to identify above-ground objects based on their size, height, and edge characteristics. First, a DSM is created from the point cloud, and then morphological gradients are calculated to analyze local elevation changes. Small objects near the ground surface are removed using a Modified White Top-Hat (MWTH) transform with directional edge constraints and a height difference threshold. This process is iterated at hierarchical levels with increasing window size to eliminate progressively larger objects.  

\Citet{slopedbrim} proposed a modification of the Top-Hat filter by introducing a sloped brim. The authors showed how this modification makes the method more robust for complex objects and terrain. Similar to the previous approach, internal and external gradients where found using a fixed structural element, and then a height-thresholded Top-Hat is used. For every (row, column) in the DSM, a pixel is determined to be ground if the difference between the original and the morphologically `opened' heights are not larger than a predefined threshold.

 \Citet{Arefi2009AUTOMATICGO} used morphological reconstruction to estimate the DTM using geodesic dilation. This process used two input images termed a  `marker' and `mask'. 
 The reconstruction procedure suppresses unwanted regions of high intensity (\ie\mbox{ }non-ground) while preserving the intensity of the regions of interest; these are seeded with `markers'.
 The marker image, generated by subtracting an offset from the DSM image, is dilated by an isotropic structuring element, while the mask acts as a limit for the dilated output. 
 After grayscale reconstruction, heights less than the offset values are removed, with the remaining heights representing the final DTM.

Alternatively, \citet{Kraus2001} proposed a method based on surface interpolation. A rough approximation of the surface is computed, and then the residuals \ie~the oriented distances from the surface to the measured points, are computed. Each point is weighted according to how far it is from the surface, and then the surface is recomputed based on these weights. A point with a high weight will attract the surface, resulting in a small residual, whereas a point with a low weight will have little influence on the surface. This process is repeated iteratively, while simultaneously classifying points as ground/non-ground based on their distance from the surface.

While the methods summarized above successfully extracted the DTM in the study areas given, they each required a set of manually-tuned threshold values to achieve the final result. While this might be tractable for a specific scene, it does not scale efficiently for large amounts of data and when no prior knowledge is given about the scene characteristics, such as the noise, the largest object size, or the maximum terrain slope. 

Recently, a new class of methods have tried to move away from the manual fine-tuning of heuristic parameters. These learning-based methods propose to estimate the DTM by constructing a binary classifier that can identify ground and non-ground data, based on previously annotated scenes. 

For example, \citet{rs8090730} used a Convolutional Neural Network (CNN) to classify pixels as ground/not ground. Since CNN architectures are generally designed to work on RGB raster imagery, the authors created training images as follows: For each point, a $(128\times128)$ image patch was generated, i.e., a DSM patch with each pixel having an intensity value corresponding to the $z$-value of that point. From each image patch, three channels were created by subtracting from the DSM patch the minimum, maximum, and mean $z$-values for that patch. The results are treated as surrogate red, green, and blue channels for a binary, deep CNN framework designed to determine whether the point belongs to the ground or non-ground class.
While the network performed well, it required a very large amount of training data (more than 17 million labeled samples). Furthermore, it involved a significant amount of overhead, since every point was represented by a $(128\times128)$ window, and the label for the window had to be transferred back to the corresponding 3D point.

Instead of classifying the input image into a ground and non-ground cover types, \citet{GevaertDTM2018} formulated the problem of estimating the DTM as a semantic labeling task where each pixel in the input RGB image is labeled (road, car, tree, etc.). To create labeled training data, the corresponding DSM image is used with a Top-Hat filter to generate labeled data. Next, a Fully Convolutional Network (FCN) \citep{FCN} framework is used for semantic labeling of the input. In order to avoid downsampling the input image, the authors suggested using dilation filters to capture the features at multiple scales. Once an image is segmented into semantic labels, the ground points are isolated and used to interpolate a DTM based on traditional methods, e.g., kriging.

Combining information from the DSM and corresponding multi-specrtal images,\citet{tapper2016extraction} trained a simple feed-forward neural network to classify a scene into ground, vegetation, and man-made objects. The input was 23 features (representing attributes such as color, texture, and geometrical descriptors) feeding into 46 hidden units then to a three outputs classifier. Using the classification labels, morphological operations is used to refine and filter ground regions. Next, the DTM is interpolated using the refined ground points. 

Finally, \citet{DTM_ENCODER} formulated the problem in an encoder-decoder framework. The authors developed a sparse autoencoder where the input is the original DSM and the output is the DTM. From this perspective, the input is treated as `noisy' data, whereas the DTM is considered to be the clean data. This approach bypasses the classification stage to predict the DTM directly from the input, similar to the approach we propose. However, the input data are limited to a scanline from the DSM, which doesn't capture the surrounding two-dimensional neighborhood information. Also, the autoencoder was constructed using a fully-connected architecture, which by design, does not capture local features.

Based on this summary of related work, we conclude that there are still several limitations with respect to extracting the DTM from point cloud data. First, traditional methods based on slope-, TIN-, or morphological filtering rely on manually-tuned heuristics and do not scale well to variable data sets. Second, most of the available methods rely on first classifying the points into ground and non-ground cover types; a potentially unnecessary step which adds complexity and computational cost. Third, learning-based methods (while having the potential to address the challenges above) are designed for 2D raster imagery, and require a costly data transformation process in order to operate on 3D point cloud data directly. Together, these limitations hinder the ability to efficiently, accurately, and objectively extract DTMs from large-scale EO point cloud data for anywhere in the world. 

To meet this need, the goal of this research is to answer the following questions: 
Given the task of extracting the DTM from a 3D-point cloud,
\begin{enumerate}
\item Can we eliminate the dependency on multiple parameters, while still incorporating spatial context, and thus reduce the time and effort spent on manual tuning?
\item Is it possible to estimate the DTM without classifying points into ground and non-ground cover types? 
\item Can such a model operate directly on the point cloud instead of relying on derivative representations created from the data, such as images or TIN models?
\end{enumerate}

\subsection{Contribution}
\label{sec:contribution}

This paper makes the following contributions, as related to the science questions (1)-(3) above.

(1) All of the existing methods agree on the importance of using context within a neighborhood to decide whether a point is ground or not. For example, slope-based methods examine neighboring points to determine a height and a slope threshold, while  morphological methods rely on the coverage area of the structural element to remove objects within a window. However, they require manual tuning of several parameters to achieve the desired results. Similarly, the  learning-based methods define a window around the point of interest to capture contextual information before passing it to the network of choice. In this paper, we extend our previous work~\citep{my_isprs_paper} to learn neighborhood information while preserving the simple and elegant architecture of the PointNet network \citep{pointnet}.

(2) Although learning-based methods introduced the application of deep learning to estimate the DTM, they typically cast the problem as a binary classification task to distinguish between ground and non-ground points. In fact, a majority of existing methods rely on first classifying the points into ground and non-ground cover types.  Such methods rely heavily on labeled points (e.g., tree, car, pavement) to train the algorithms. Here we design a learning-based method that bypasses the need for classifying ground points, thus eliminating the need for labeled training data. Instead, our model is trained on truth DTMs directly, and can subsequently be easily deployed on any point cloud for inference. 

(3) Learning-based methods operate on a transformed version of the point cloud \ie~DSM images, in order to meet the requirements of deep learning architectures that were designed for 2D imagery. This transformation introduces unnecessary overhead and may introduce a loss of information. For example, if a DSM image is created from LiDAR points, one has to decide which return to use at a given pixel, as well as finding a way to fill in the void regions where data are missing. Similarly, when using optical photogrammetry, multiple points may exist at a given pixel; this requires a decision on which points are used in constructing the DSM. Here, we operate on the input point clouds directly, therefore respecting their unstructured and unordered nature.



In summary, inspired by the success of methods that operate directly on point clouds and allow the seamless integration of geometric and spectral information~\citep{my_isprs_paper,pointnet}, we make the following contributions: 
\begin{itemize}
\item We design a learning-based method that bypasses the need for classifying ground points.
\item Our algorithm predicts the DTM value for each input point within the 3D point cloud without the need for intermediate steps \eg~creating a TIN model.
\item We operate on the input point clouds directly, therefore respecting their unstructured and unordered nature.
\item We extend PointNet to learn neighborhood information while keeping the simple and elegant architecture of the network.
\item We eliminate the need for calculating handcrafted features derived from the DSM or fine tuning parameters that are scene-specific, \ie~Top-Hat's structural element.  
\end{itemize}

This paper is organized as follows: Section 3 will describe the network used and the extension of PointNet to handle neighborhood information. Sections 4 will present the evaluation of our approach. Finally, Section 5 draws the conclusions and presents our vision for future work.

\section{Methodology}
\label{Methodology}
In this section we present a deep learning method that predicts the DTM value for each point in an input point cloud. Recognizing the importance of spatial context, our method is able to learn point-level, neighborhood-level, and global features directly in an end-to-end fashion, rather than relying upon handcrafted features and hand-tuned parameters. The term `global' in our case, represents the block-wise (\ie~the input set) features, unless stated otherwise in the text.

This end-to-end learning operates directly on the point cloud, without relying on derivative image-based representations (contribution 3). 
We do this by extending the PointNet architecture.
Although one could design a simple, fully connected network that accepts a single point at a time, such a design would be limited in that it would only learn a single point-level feature for each point.  
Instead, PointNet proposed an effective, yet very simple, deep learning architecture that consumes `blocks' of point cloud data and aggregates point-wise features into a global feature vector that represents the entire input point set (\ie~block-wise)  features. This work showed promising results when applied to tasks such as semantic labeling of aerial LiDAR data~\citep{my_isprs_paper}. 

However, the aggregation of point-wise features to create a global representation for the input block set suggests that the network doesn't capture the neighborhood contextual information. Several attempts have been made to address this issue. For example, \citet{PointNet++} proposed PointNet++ where they group a point set into smaller clusters, then apply PointNet recursively on a nested partitioning of the input set. Unfortunately, this modification to the architecture fails to preserve the point-wise features in their object classification architecture and the global features in their segmentation network. 

In an attempt to retain the point-wise and global features of PointNet, \citet{Shen_Kernel_2018} proposed the use of a kernel correlation layer to exploit local
geometric structures using a graph-based approach. They construct K-nearest neighbor graphs (KNNG) by considering each point
as a vertex, with edges connecting only nearby vertices. Using the graph information, kernels were learned and applied to the vertices to capture local information and concatenated with the point-wise and the global vectors from the original PointNet network. However, this method introduced unnecessary complexity and computational overhead. 

\begin{figure*}[tp]
\begin{center}
\includegraphics[width=13cm]{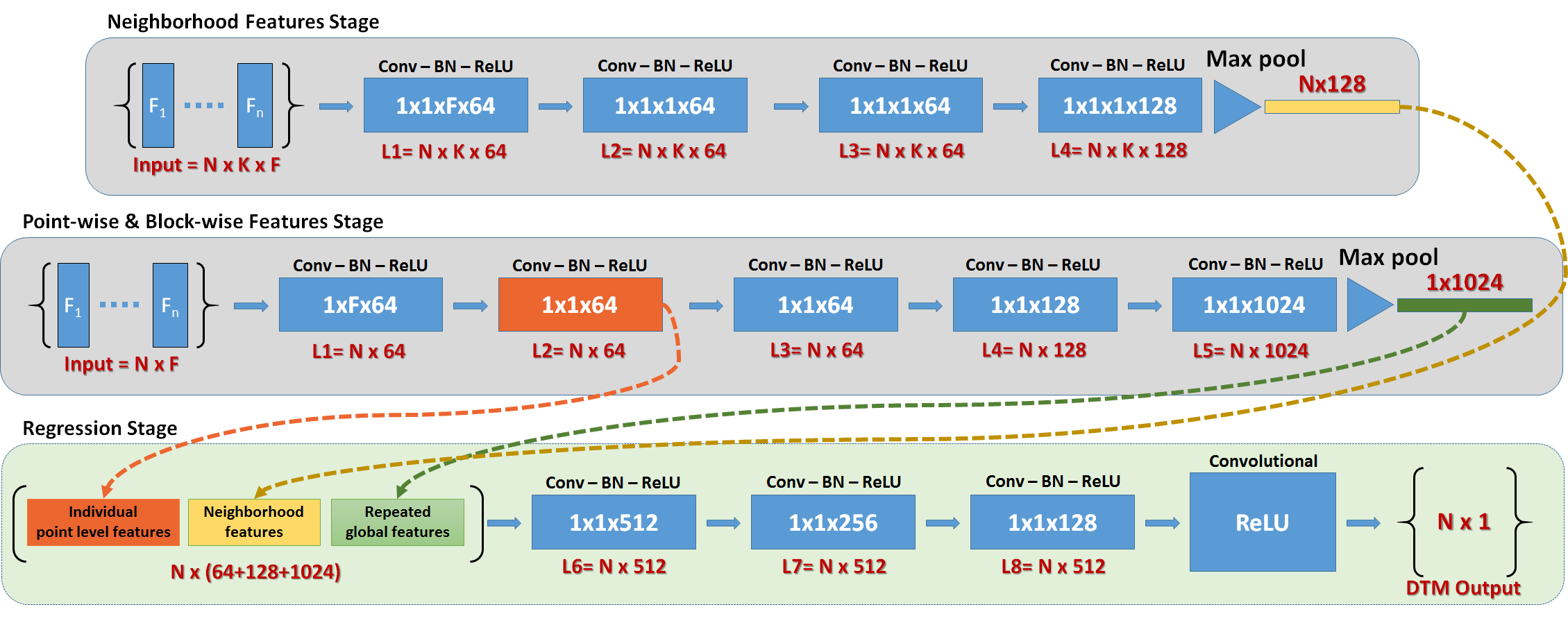}
\end{center}
\caption{
The network takes as input two representations of the points within a block. The first representation (center panel) is a set of size $N\times~F$ used to learn point-wise features and later pooled to generate a global features for that block. The second set (upper panel) is of size $N\times~K\times~F$ and is used to learn neighborhood features within that block, where K is the number of neighborhood points (see~\ref{subsection:Neighborhood}).
Next, features learned from both sets are concatenated and passed though $(1\times1)$ convolutional layers and then to a regressor to perform DTM prediction per point (lower panel), \ie~the $N\times1$ output. This can be repeated for each block within the point cloud. Text in white indicates the filter size, while text in red indicates the layer's output shape.}
\label{fig:algorithm}
\end{figure*}

Instead, we extend the original PointNet to allow the use of the point-wise, neighborhood, and global features, while retaining its simple, yet effective  architecture (contribution 1). The complete architecture of our network is shown in Figure \ref{fig:algorithm}.
\subsection{Network Architecture}
Our network takes as input \emph{a pair} of blocks. The first is a block $X$ of $N$ 3D-points, where $X$ = \{$x_{1},x_{2},x_{3},...,x_{N}$\}. The second set, however, is a $N\times K\times F$ array representing the neighborhood data within that block. Note that this is merely an alternative structuring of $X$, just grouped and replicated in such a way as to provide spatial context. Further details on the neighborhood block will be explained in Section~\ref{subsection:Neighborhood}.

In addition to spatial coordinates, the points in block $X$ potentially have point-level feature attributes such as return number, intensity and/or spectral information, such that $x_i$ is of length $F$  where $F$ is the number of feature attributes per point, e.g., $x_i = {\textsc{x}}_i,{\textsc{y}}_i,{\textsc{z}}_i,{\textsc{r}}_i,{\textsc{g}}_i,{\textsc{b}}_i$. 
Given the truth terrain height $Y$ = \{$y_{1},y_{2},y_{3},...,y_{N}$\} for the training set,
the goal of DTM estimation is to predict the $z$-value for the ground-level at every point in the test set, \ie~$\hat{Y} = \{\hat{y}_{1},\hat{y}_{2},\hat{y}_{3},...,\hat{y}_{N}\}$. For each input set, the network ingests this $N\times F$ array of unordered data points $X$, (along with an $N\times K\times F$ array representing the neighborhood data) and outputs a $N\times1$ array $\hat{Y}_i$ where each entry is the DTM prediction per input point. This can be repeated for each input set (\ie~ a block of data), with the results aggregated so as to return the DTM estimates for each point in the entire point cloud. For a given scene, we subdivide our training data into overlapping blocks, thus increasing the number of training samples.

 Although the network is designed to flexibly accept inputs with different numbers $F$ of features, we demonstrate the network using the following 10 features given the available data: Each point $x_i$ is represented as a 10D-vector, containing the block-wise centered coordinates ($\textsc{x},\textsc{y},\textsc{z}$) representing the normalized coordinates with respect to the input block, spectral data ($\textsc{ir,r,g}$), the Normalized Difference Vegetation Index~\citep{NDVI} ($\textsc{ndvi}$), and the normalized coordinates ($\bar{\textsc{x}},\bar{\textsc{y}},\bar{\textsc{z}}$) with respect the full extent of the scene. 



To learn point-wise features, the $N\times F$ array of points from each block of data is convolved with a set of 1D-filters of size $1\times F$, \ie~the filters are applied across the the columns to capture the interactions between the point attributes as shown in the central portion of Figure~\ref{fig:algorithm}. It should be noted that this operation can be carried out using a 2D-filter while making sure that the filter is applied only in one dimension. This allows us to define 1D-filters in the form on ND-filters, thus allowing a simpler implementation, by ensuring that the convolution operation is applied on a specific axis. The output of this operation is an $N\times F'$ array with $F'$ representing the new point-wise features. To capture an embedding that describes the whole input set, \ie~the block, the point-wise features are aggregated using a pooling operation (\eg~maxpooling, average pooling, or weighted average pooling) to a single vector.
In this stage (center panel in Figure \ref{fig:algorithm}), the network can only learn point-wise features and a global feature vector per input block. Note the similarity to PointNet. 
\subsection{Learning Neighborhood Features}
\label{subsection:Neighborhood}
It was pointed out in Section \ref{sec:related_work} that neighborhood relationships are critical in estimating the DTM value at a given point \citep{rs8090730,MWTH2014}. To allow our network to capture neighborhood information, we need to first determine which points are within a `neighborhood'. Several options exist. Using a Kd-tree search, one can either find the closest $K$ points to the query, or define a search area and collect all points within that area. In the first case, while the number of neighborhood points is fixed, the points themselves may be found at varying distances from the query point. Alternatively, while a predefined search area allows for neighborhoods at different scales, the number of points found within a fixed search area may be different from one point to another.

Recognizing these trade-offs, we take the following approach: For every point $x_{i}$ in the input set where $i$ = \{$1,2,3,...,N$\}, we define the neighborhood as the set of $K$ sampled points found within a sphere centered at $x_{i}$ with a radius $R$. 
 This ensures that the total number of points within a fixed radius will be the same for all points; we replicate points if the available number of samples is less than $K$.

In this stage (upper panel in Figure \ref{fig:algorithm}), each individual input point within the block can be represented by its neighborhood points in the form of an $N\times K\times F$ array. Recall that $N$ is the number of points within the input block, $K$ is the number of neighborhood points, and $F$ is the number of attributes per point within the neighborhood. 
Note that in PointNet++, the algorithm first subsets the original input data before finding the neighborhood; this results in an array of $N'\times K\times F$ where $N'<N$. However, as opposed to classical computer vision tasks such as object classification, where fewer points can describe an object, estimating the DTM is done at every point in the scene. Therefore, we choose not to exclude points from the input data set.

Next, the $N\times K\times F$ input array is convolved with a set of $F'$ 1D-filters along the feature dimension, \ie~a single filter shape is $1\times 1\times F$ as shown in Equation \ref{eq:1},
\begin{equation}
\label{eq:1}
G_{(N,K,F')} = X_{(N,K,F)} * h_{(1x1xF)}
\end{equation}
where G is the output of the convolution process with a shape of $N\times K\times F'$, and $F'$ is the number of new features learned (\ie~embeddings) for every point within the neighborhood. At this stage, every point in the input array is still represented by the corresponding features embedded in the collection of $K$ neighboring points. 

In order to summarize this neighborhood information for every input point, we aggregate/pool the information from all neighboring points to form a single vector per-point that encodes the neighborhood information as follows:
\begin{equation}
\label{eq:2}
G' = max~(0,G,axis=1)
\end{equation}
where $G'$ is an array of size $N\times F'$ and each input point is now represented by the encoding of all the features within the neighborhood using the $max$ operation along the neighborhood axis.
The result from Equation~\ref{eq:2} is then concatenated with both the point-wise features and the global feature vector from the original PointNet architecture as show in Equation~\ref{eq:3}
\begin{equation}
\label{eq:3}
T = concat~(P,B,G')
\end{equation}
where $P$ is the point-wise feature vector, $B$ is the global feature vector, and $G'$ is the neighborhood vector. These are represented in Figure \ref{fig:algorithm} as the orange, yellow, and green boxes, respectively. 

\begin{figure*}[!htbp]
\begin{center}
\includegraphics[width=13cm]{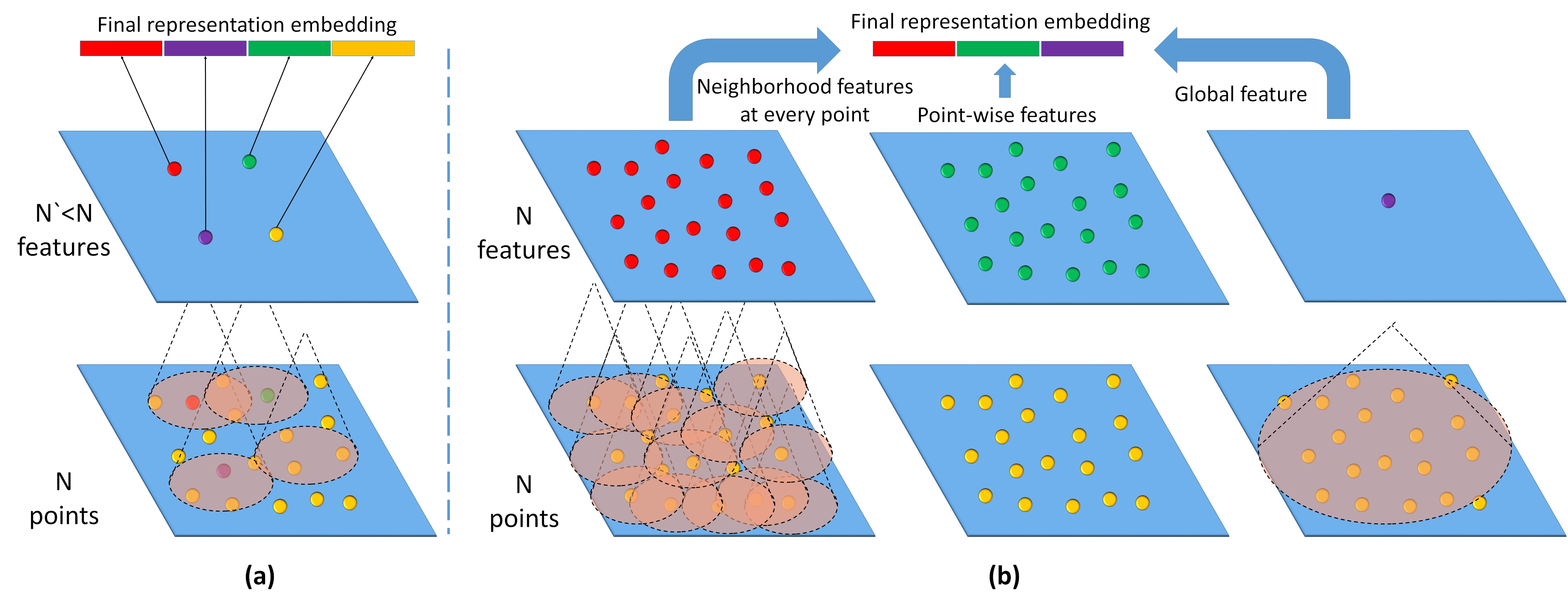}
\end{center}
\caption{The figure shows the difference between PointNet++ (a), and our network (b), in terms of the final feature representation. Our network combines all point-wise, neighborhood, and global features from all input points, while PointNet++ only aggregates the information learned from a sampled set neighborhood to create a final representation. The blue tile represents a block.}
\label{fig:Compare}
\end{figure*}

It should be noted that PointNet++ aggregates the information from the neighborhood data only to produce the final representation without including the point-wise and the global feature vectors found using the original PointNet architecture. 
Figure \ref{fig:Compare} shows the difference between our architecture and PointNet++ presented by \citet{PointNet++}. As shown, we learn neighborhood information for every point within the block, instead of for just a sampled set of the points from the block. In addition, we learn point-wise and global features.    

\subsection{Inference}
\label{Inference}
The output from Equation~\ref{eq:3} is passed through a series of a $1\times 1$ convolution layers and finally to a linear layer with a ReLU activation as shown in Figure~\ref{fig:algorithm} (bottom panel). The ReLU activation forces the outputs to be greater than or equal to zero. This guarantees that all the DTM estimations are positive. This design can be thought of as solving the following constrained optimization task:
\begin{equation}
\begin{aligned}
\label{eq:4}
& \underset{w}{\text{minimize}}
& & cost~(Y_{i},f_{i}(X;w)) \\
& \text{subject to}
& & f_{i}(X;w) \geq 0
\end{aligned}
\end{equation}
where $Y_{i}$ is the ground truth values, $f_{i}(X;w)$ is the network prediction, $X$ is the input array, and $w$ are the network parameters to be learned. 

To train the network, two popular loss functions can be considered; Mean Squared Error (MSE) and Mean Absolute Error (MAE). An advantage of using a MSE loss is the convex nature of the function, which makes is easier to optimize with linear derivatives. However, since the errors are squared, it is very sensitive to outliers, which are common in LiDAR data due to multiple scattering, or in optical point clouds due to imperfections in dense matching (\eg~ spikes) from multi-view imagery. To avoid biasing the prediction based on these outliers, while ignoring other good samples, one can use the MAE loss function. However, since this loss function uses the absolute value operator, it is not differentiable at zero, and the derivatives are constant on both sides of the origin due to the linear nature of the function.
Seeking a trade-off between these loss functions, we choose the logarithm of the hyperbolic cosine function as shown in Equation~\ref{eq:5}, which offers a balance between the MSE and MAE behaviors. 
\begin{equation}
\label{eq:5}
\ell = \sum_{i=1}^{N} \log(\cosh{(Y_{i} - \hat Y_{i})})
\end{equation}
It is a convex function with a quadratic behavior at small values, allowing for non-constant derivatives in this range, while providing a linear behavior at large values. This makes it insensitive to outliers while remaining fully differentiable.

\section{Evaluation}
\label{sec:evaluation}
\subsection{Dataset}
\label{sec:dataset}

For this paper, we use two different sources of point cloud data, comprising 4 individual test sets. The first set is LiDAR data provided by the ISPRS 3D-Semantic Labeling Contest, as part of the urban classification and 3D-reconstruction benchmark\footnote{\url{https://goo.gl/6iTj6W}}. The remaining three sets are sample point clouds derived from dense stereo matching using multiple-view imagery available from DigitalGlobe and processed by Vricon \citep{tripoli-data}. 

The LiDAR data were acquired using a Leica ALS50 laser scanner flown at a mean height of 500m above ground level over Vaihingen, Germany. This resulted in a point density of approximately 4 points/m$^{2}$. 
Additionally, corresponding georeferenced IR-R-G imagery was provided, with a ground sampling distance of 8cm and dimensions of $20250\times 21300$ pixels.
We used the image data to extract the spectral attributes for each point in the LiDAR data.

The LiDAR coverage area was subdivided into two regions; one for training and the other for testing, with a total of 753,859 and 411,721 points, respectively. 
The training area is comprised mostly of residential detached houses. It covers an area of $399\mbox{m}\times 421\mbox{m}$.
The test area, on the other hand, is located at the center of the Vaihingen city proper, and is representative of complex urban structures. It covers an areal extent of $389m\times 419m$.

The dense optical point clouds were generated by Vricon, which employs a dense reconstruction from multiple-view satellite imagery \citep{vricon, vricon2}. High-resolution, multispectral imagery from DigitalGlobe's constellation of WorldView satellites \citep{anderson2012worldview} provides a nominal post spacing of 0.5m with spectral features in the R,G,B, and IR bands. 

Three scenes were used, all based in in Tripoli, Libya. The first scene, of size $1000\mbox{m}\times 1300\mbox{m}$, contains 3.7 million points within an extra-urban area, with many high-rise buildings distributed about a large open area.
The second scene, of size $1500\mbox{m}\times 500\mbox{m}$ contains 3 million points and is representative of smaller urban structures with wide roadways. The last scene covers the historic old town region of the city, with densely-packed structures and very little ground visible except in sparsely distributed areas. This scene covers an area of $650\mbox{m}\times 450\mbox{m}$ with 1.1 million points. All of the image-derived point clouds have a point density of 4 points/m$^2$. 

\begin{figure*}[!htbp]
\begin{center}
\includegraphics[width=14cm]{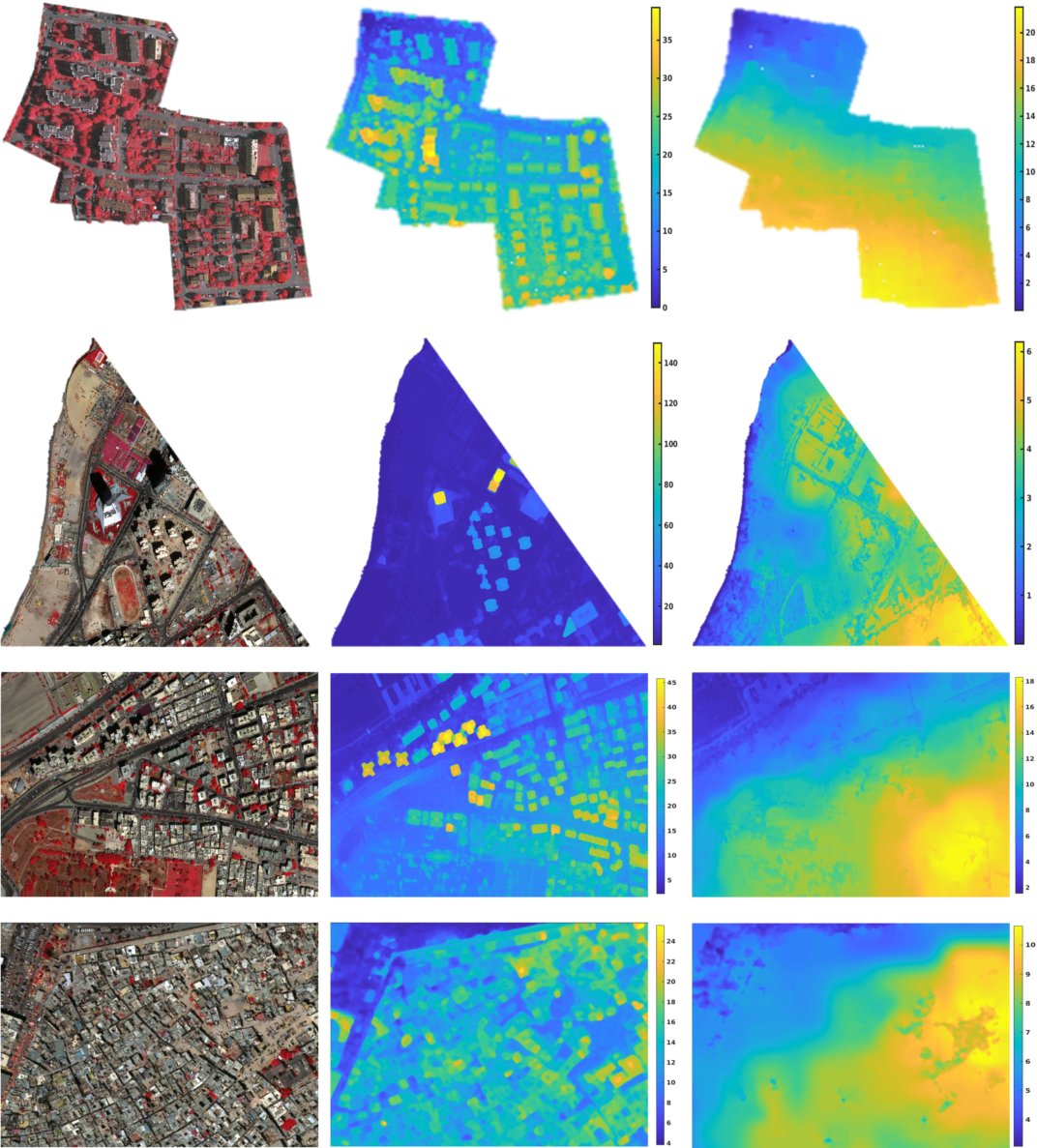}
\end{center}
\caption{Training data sets, from top to bottom: ISPRS LiDAR data, Vricon high-rise area, lower extra-urban data, and historic old-city area. For each, we render, from left to right: point cloud colored using the spectral attributes, the DSM (showing height information), and the ground-truth DTM.}
\label{fig:dataset_train}
\end{figure*}

\begin{figure*}[!htbp]
\begin{center}
\includegraphics[width=14cm]{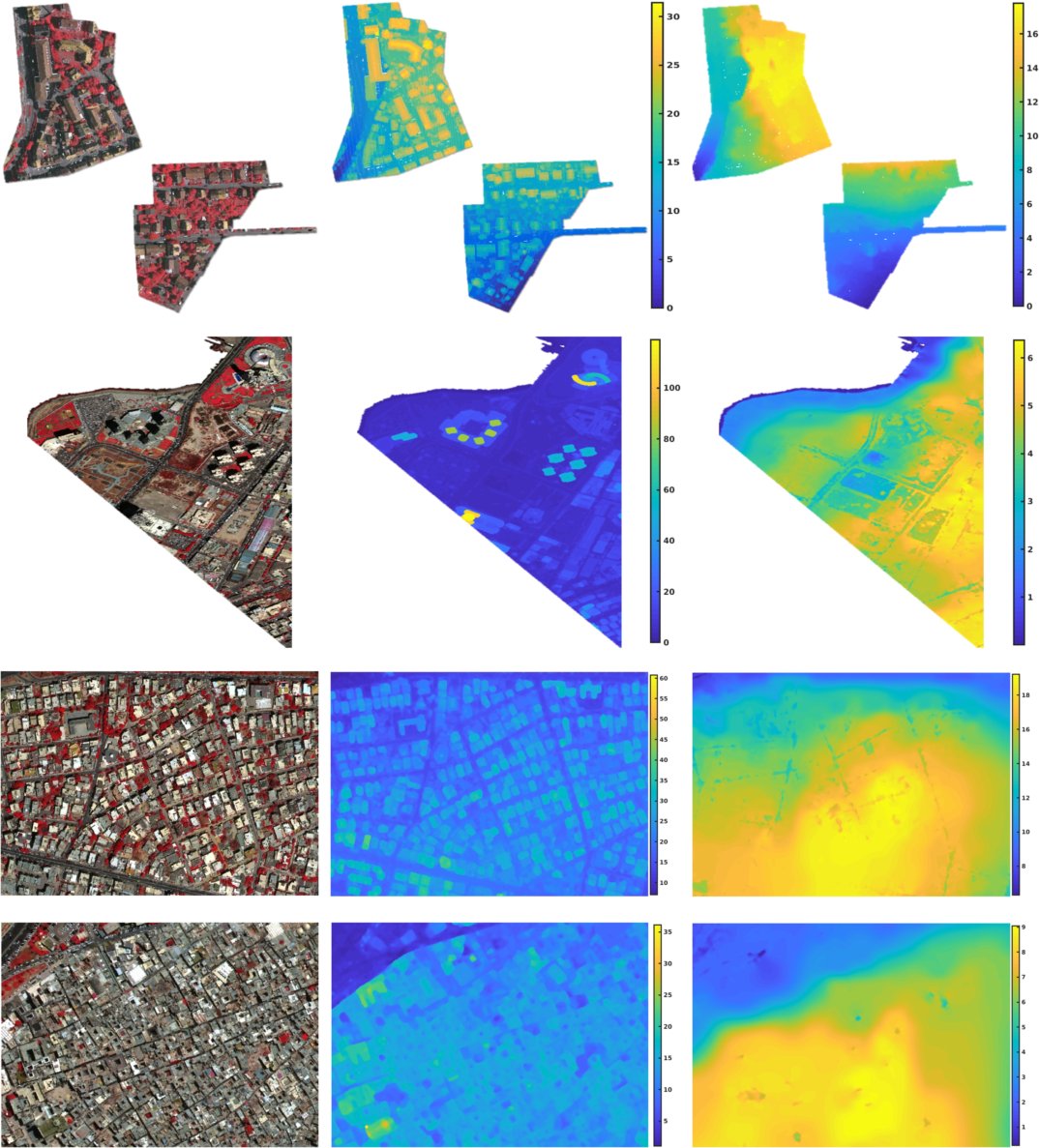}
\end{center}
\caption{Testing data sets, from top to bottom: ISPRS LiDAR data, Vricon high-rise area, lower extra-urban data, and historic old-city area. For each, we render, from left to right: point cloud colored using the spectral attributes, the DSM (showing height information), and the ground-truth DTM.}
\label{fig:dataset_test}
\end{figure*}

For the purposes of this study, we have subdivided these scenes into training and testing areas, each containing 50\text{\%} of the data. 
Figure~\ref{fig:dataset_train} shows the training data set including point clouds and their corresponding DSM and ground-truth DTM for all four scenes.  Figure~\ref{fig:dataset_test}, likewise, shows the testing data set.

The ground truth data were obtained as follows: For the ISPRS data, we use the DTM generated by LAStools as our ground reference. For the optical image-derived data, we use a DTM of 0.5m resolution, provided by Vricon.
We acknowledge that there may be errors in both DTM sources, in particular for the LAStools output which is not subject to production-level quality control. However, both data sources add unique value to this study: The ISPRS data are used due to its recent popularity as a benchmark for the semantic labeling tasks, allowing us to share our findings on a recent publicly available dataset. And the Vricon data are used to evaluate the network on 3D data obtained using optical photogrammetry, thus demonstrating the feasibility for large-scale DTM extraction from EO satellites. 

\subsection{Preprocessing}
\label{sec:preprocessing}

During training and testing, we represent each point by its 10D-vector ($\textsc{x}$, $\textsc{y}$, $\textsc{z}$, $\textsc{ir}$, $\textsc{r}$, $\textsc{g}$, $\textsc{ndvi}$, $\bar{\textsc{x}}$, $\bar{\textsc{y}}$, $\bar{\textsc{z}}$) (see Section~\ref{Methodology} and Figure~\ref{fig:algorithm} for details). For the LiDAR data, the spectral information is extracted by interpolating the georeferenced IR-R-G imagery at the point locations for the LiDAR-derived point cloud. For the optical point clouds extracted from multiple-view imagery, this spectral information is inherently defined for each point. 

The validation data are created by sampling  the training data. The training data are augmented by randomly rotating the points to alter their geographic orientation around the $z$-axis. We also add random noise to the coordinates after each rotation. The additive noise is sampled from a zero-mean Gaussian distribution with $\sigma=0.08$m for the $x,y$ coordinates and $\sigma=0.04$m for the $z$ coordinates. These values were chosen empirically such that the overall distribution of the data are similar while adding sufficient noise.
The original training data are discarded and only the augmented data are retained to avoid similarity between the training and validation data. 

Since deep learning models require large amounts of training data, we subdivide the scenes into smaller blocks with 50\text{\%} overlap. \Citet{my_isprs_paper} showed that the size of the blocks affects the classification quality due to the effect of the global feature learned from each block. They used $2m\times 2\mbox{m}$, $5\mbox{m}\times 5\mbox{m}$, and $10\mbox{m}\times 10\mbox{m}$ blocks to capture objects at small, medium, and large scales. However, in the task of estimating the DTM, we are interested in a more synoptic view of the scene to capture the changes in the terrain. Therefore, the $10\mbox{m}\times 10\mbox{m}$ block size was chosen for all experiments. 
During training, we instantiated each block with $N$=2,048 points sampled from within each block for computational purposes. Note that we operate on all 2,048 points, while PointNet++ selects a smaller subset out of this input to learn the neighborhood information. During testing, we use all the points within a block.
By fixing the block size, we are limiting the network to learn only from the geographic area delineated by that fixed-size block. To overcome this issue, our network is designed to accept neighborhood information that complements the point-wise and the global features learned from the blocks. The user has the flexibility to choose the neighborhood radius, depending on the task. 
\subsection{Training Parameters}
\label{sec:training_parameters}
To generate the $N\times K \times F$ array containing neighborhood relationships, we define a search radius $R$=25m and sample $K$=128 points from within each spherical region, thus creating an array of dimensions $2,048\times 128\times 10$ where each of the original 2,048 points within a block is represented by 128 points from its 25m neighborhood, each with a 10-dimensional feature vector. While sampling 128 points from the 25m radius area seems very small, applying this for every point within the block sufficiently covers the neighborhood when viewed together.

During the training process, we use the Adam optimizer~\citep{Kingma2014AdamAM} with an initial learning rate 0.001 and a momentum of 0.9. Since the inputs to the network are a $2048\times 10$ array and a $2048\times 128\times 10$ array, we choose a batch size of 12 due to memory limitations in our current computing environment.
The learning rate is linearly reduced after each iteration as a function of the current number of epochs. We train the network for a total of 16 hours for 20 epochs and monitor the progress of the validation loss to save the weights as the loss improves. 
We train using an NVIDIA Tesla P40 GPU and Keras~\citep{keras} with the Tensorflow backend.
\subsection{DTM Results}
\label{sec:DTM_results}
\begin{figure*}[!htbp]
\begin{center}
\includegraphics[width=13cm]{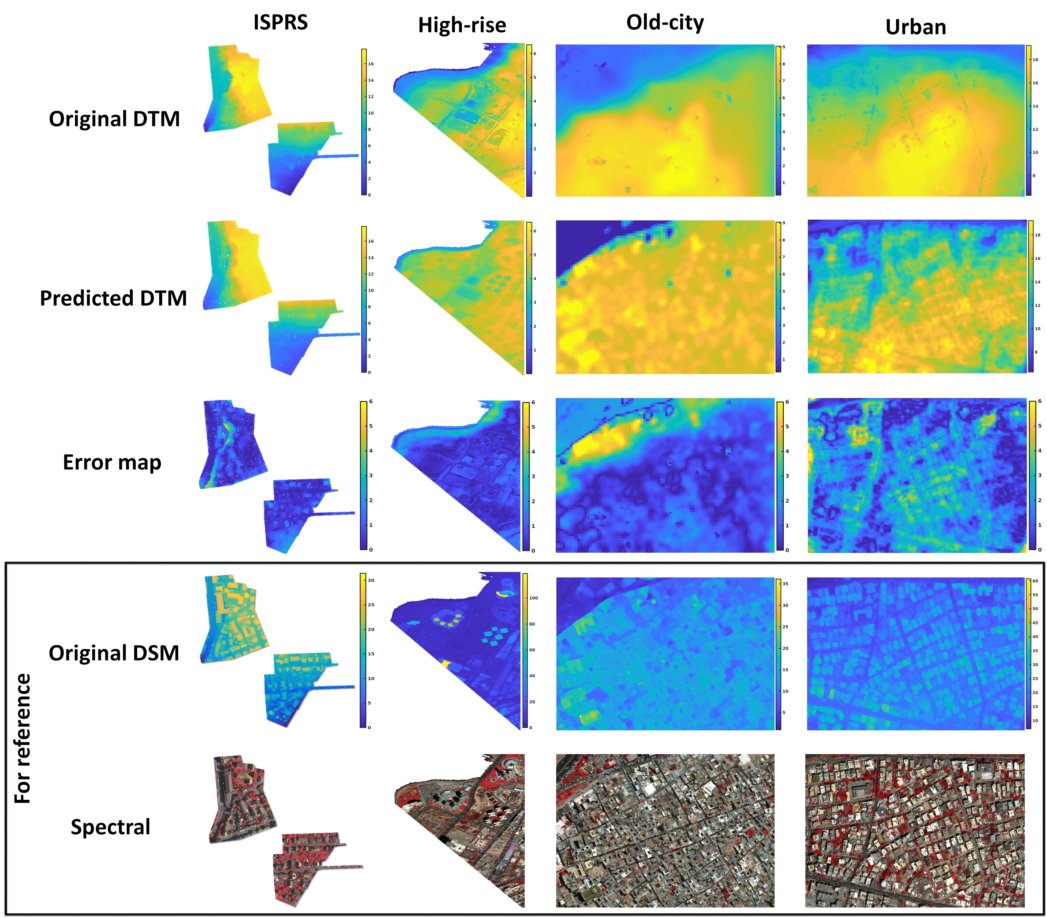}
\end{center}
\caption{For each of the four testing scenes, we show the estimated DTM $\hat{Y}_i$ in comparison to ground-truth DTM $Y$. The difference gives an error map, $|Y-\hat{Y}|$. The original DSM and colorized point cloud is shown in the bottom panel for reference.}
\label{fig:results-overview}
\end{figure*}

For each of the four testing scenes, we estimate the DTM $\hat{Y}_i$ at every $i^{th}$ point within the scene and compare that to ground-truth data $Y$, to generate an error map, $|Y-\hat{Y}|$. During testing, we subdivide the data into $10\mbox{m}\times 10\mbox{m}$ blocks, similar to training, and report the DTM estimate for all input points. These summary results are shown in Figure~\ref{fig:results-overview}; the bottom panel shows the original DSM and colorized point cloud, for reference. Note that these results are for a neighborhood radius of 25m. More information on the impact of the neighborhood radius is presented in Section \ref{sec:effect_of_neighborhood_size}. It should be noted that our results were reported without any post-processing,~\ie without smoothing the predicted DTM. Although post-processing could improve the results, our intent here is to present our findings in their original form.

From these error maps, we calculate the mean absolute error (MAE) and the standard deviation $\sigma$ of the error between the ground truth DTM values $Y$ and the predictions obtained from our network $\hat{Y}$. Quantitative results are summarized in the right-most column of Table~\ref{tab:results}.

Furthermore, to benchmark our DTM extraction method against alternative solutions, we also compute the MAE and $\sigma$ values for the results obtained from two widely-used software packages for DTM extraction from point clouds, namely ENVI~\citep{ENVI} and LAStools~\citep{LAStools}. 
These results are presented in the center columns of Table~\ref{tab:results}.
Note that this comparison is only done for the Tripoli scenes with Vricon ground truth to avoid a circular comparison (the only available ground truth for the LiDAR dataset was also from LAStools; see Section \ref{sec:dataset}). 
So although MAE and $\sigma$ values are not provided for the LiDAR dataset, in Section \ref{sec:effect_of_neighborhood_size} we will show how the LiDAR dataset allows us to examine the effect of the neighborhood size.


A note on the settings we used when evaluating the software packages: Both LAStools and ENVI present the user with multiple tunable parameters. 
For example, with LAStools, various parameters are required, including `step' (\ie~wilderness, nature, town or flats, city, and metropolis), `bulge' (how high the initial DTM estimate is allowed to bulge upwards during the refining stage),  ground point search (controlling the quality of the initial ground points estimation, and represented by values such `fine, hyper, and ultra`), and `spike' (determining how prominent a spike can extend). 
Similarly, ENVI 
presents the user with multiple parameters, the three most important being: grid resolution, point filtering, and the variable sensitivity. 
We want to compare our method against the best possible results from these software packages. Therefore, we ran multiple parameter combinations in order to find the best-performing parameters for each set of training data, and then used these best-performing parameters on the corresponding test data. 

Visual results comparing our method to LAStools and ENVI are shown in Figures \ref{fig:Dense_Results}, \ref{fig:City_Results_v2}, and \ref{fig:Highrise_Results} for the old-city, urban and high-rise areas of Tripoli, respectively. Note that we have matched the color scale of all predicted DTMs to the ground truth DTM; although this saturates any predicted values that are beyond the maximum of the ground truth, it allows us to visually compare the changes with a common reference.



\begin{table*}[!htbp]
\centering
\caption{MAE in meters and the corresponding standard deviation ($\sigma$) of the error between the ground truth DTM values ($Y$) and the predictions obtained from our network ($\hat{Y}$).}
\label{tab:results}
\small
\tabcolsep=0.3cm
\begin{tabular}{c|c|c|c|c}
    \textbf{Scene} & \textbf{Metrics} &  \textbf{LAStools} & \textbf{ENVI} & \textbf{Ours (25m)}
    \\ \cline{1-5}
\multicolumn{1}{c|}{\textbf{Old-city}}   & \textbf{MAE}      & 2.00 & 3.90 &  1.17 \\
\multicolumn{1}{c|}{\textbf{}}   & \textbf{$\sigma$}       & 1.37 & 1.8 & 1.2 \\ \hline
\multicolumn{1}{c|}{\textbf{Urban}}   & \textbf{MAE}       & 1.80 & 2.0 &  1.91 \\
\multicolumn{1}{c|}{\textbf{}}   & \textbf{$\sigma$}        & 1.23 & 1.4 & 1.1 \\ \hline
\multicolumn{1}{c|}{\textbf{High-rise}}   & \textbf{MAE}  & 1.22 & 2.42 & 0.77 \\
\multicolumn{1}{c|}{\textbf{}}   & \textbf{$\sigma$}   & 1.1 & 3.02 & 0.72 \\ \hline
\multicolumn{1}{c|}{\textbf{ISPRS}}   & \textbf{MAE}  & - & - & 0.70 \\
\multicolumn{1}{c|}{\textbf{}}   & \textbf{$\sigma$}   & - & - & 0.75 \\ \hline
\end{tabular}
\end{table*}
\section{Discussion}
\begin{figure*}[tp]
\begin{center}
\includegraphics[angle=90,width=11cm]{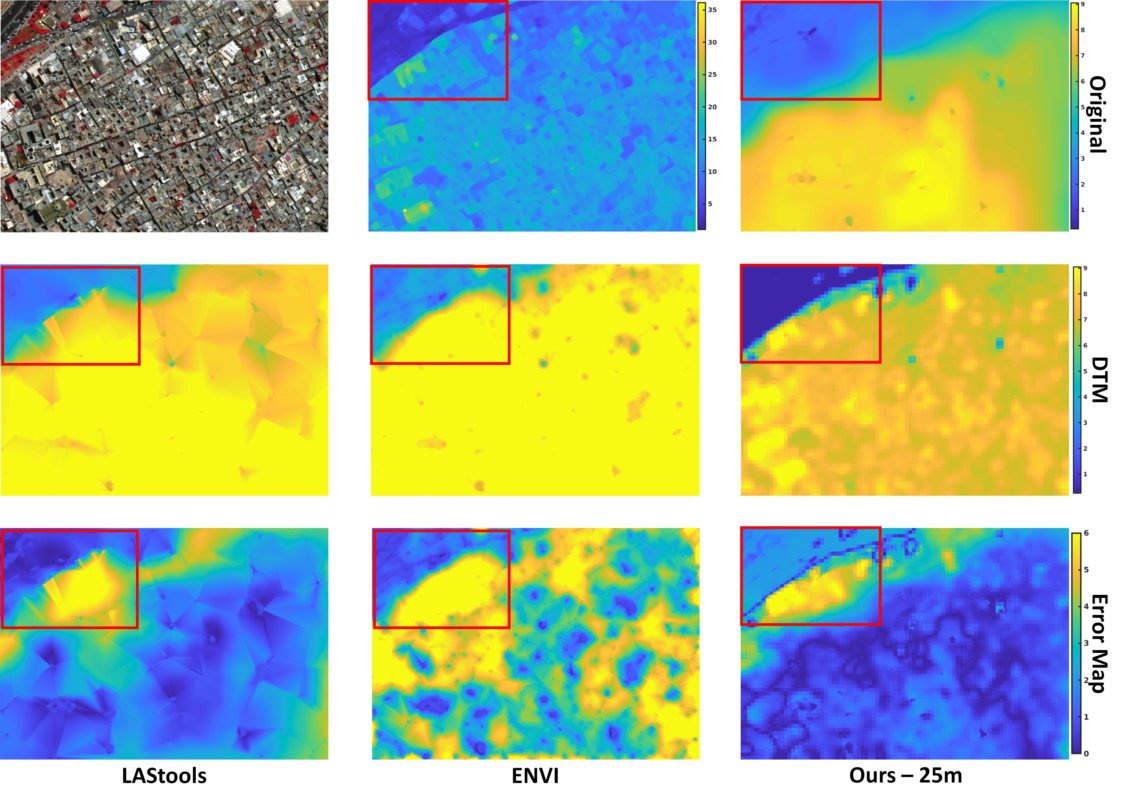}
\end{center}
\caption{Comparison of results for the Tripoli old-city scene. The first row, from left to right, shows the original color point cloud and the corresponding DSM and ground truth DTM. The second row shows the estimated DTM obtained using LAStools, ENVI, and our method. By subtracting these from the ground truth (top-right) we obtain the corresponding error maps for each method (bottom row).}
\label{fig:Dense_Results}
\end{figure*}

For the old-city scene, the best parameters in LAStools were obtained using `city or warehouses', with `bulge' is set to False, ground point search set to `hyper', and a 0.2m `spike'. For ENVI, the best results were obtained using the default parameters with the exception of setting the variable sensitivity option to `flat'. 
As shown in Table~\ref{tab:results}, our method reported a MAE of 1.17m, outperforming LAStools and ENVI, which obtained results of 2.0m and 3.9m, respectively. This is a promising result given the challenging scene content, \ie~there are very few regions in the scene where the bare-ground is visible, due to the densely-packed structures. Methods that rely on slope- or TIN-relationships, and on ground vs. non-ground classification typically over-estimate the DTM in this sort of scene, due to the likelihood of misclassification  and the lack of abrupt surface discontinuities.  These  blunders are most apparent in the ENVI prediction. 

Also, it is clear that both methods rely heavily on local minima within the scene (see the error maps in Figure \ref{fig:Dense_Results}). This results in triangulation artifacts during TIN creation. This is most apparent in the LAStools error map where the final DTM is created by `growing out' from local minima in the form of triangles. This has the effect of creating ramp-like triangles, which may seem like a smoothing effect, but actually introduce fake slopes that potentially tilt flat surfaces (\ie~ rooftops). 

On the other hand, our method bypasses the classification stage and predicts a per-point DTM value utilizing the neighborhood information. Therefore, our predictions and their corresponding errors are localized (\ie~point-wise), instead of being generated due to surface interpolation. This advantage allows us to achieve the lowest standard deviation in this scene of 1.2m compared to 1.37 and 1.8 for LAStools and ENVI respectively. 

Because of this point-wise approach, we see different behavior in the top-left region of the scene (marked with a red box). In particular, we see that our approach predicts a sharp height transition, while LAStools and ENVI predict a slowly-varying elevation profile due to their use of surface interpolation. However, this smooth transition in the ground truth DTM can not be explained by manually investigating the input DSM: 
The lowest DSM height values in the residential area adjacent to the roadside are over 7m high; we have confirmed that these are true ground surfaces by manually examining the DSM and spectral imagery. However, the ground truth DTM as prepared by Vricon estimates a slowly-varying elevation profile, with height values of ~2-3m in this same area. While it is not clear how this ground truth DTM was generated, this difference explains the behavior of our method in this region. This observation also explains the large error for all methods at the transition area. 

\begin{figure*}[tp]
\begin{center}
\includegraphics[angle=90,width=11cm]{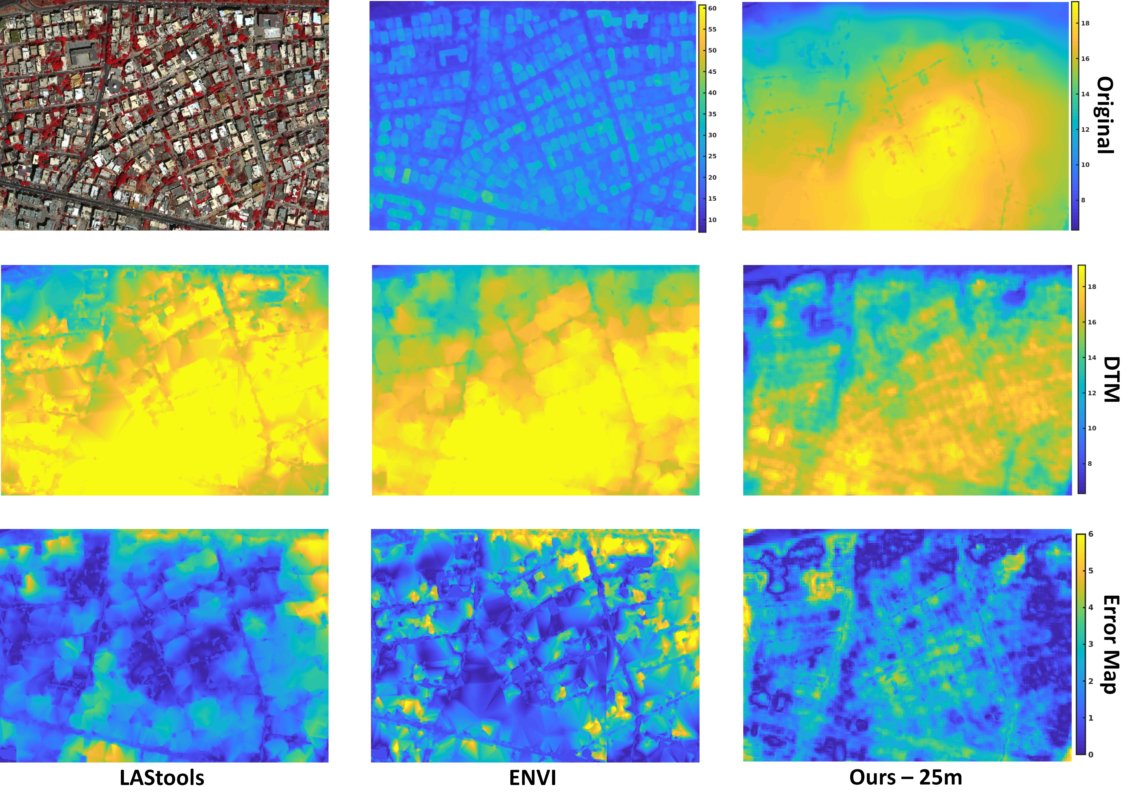}
\end{center}
\caption{Comparison of results for the Tripoli urban scene. The first row, from left to right, shows the original color point cloud and the corresponding DSM and ground truth DTM. The second row shows the estimated DTM obtained using LAStools, ENVI, and our method. By subtracting these from the ground truth (top-right) we obtain the corresponding error maps for each method (bottom row).}
\label{fig:City_Results_v2}
\end{figure*}


For the urban scene (Figure \ref{fig:City_Results_v2}), the scores for all three methods were relatively close to each other. Our method achieved a MAE of 1.91m, only 11cm higher than LAStools with a MAE of 1.8m. ENVI, on the other hand, scored a MAE of 2.0m, as seen in the second row of Table~\ref{tab:results}. For LAStools, we obtained the best results using `town or flats' without bulge and a 0.2m `spike'. Similarly, for ENVI we used a 1m grid with urban filtering, and with the variable sensitivity option turned off. 

The results from both ENVI and LAStools exhibited piecewise-linear predictions. This effect is more visible as `patches' in the ENVI DTM result due to TIN interpolation. Note that some areas are saturated due to the consistent color scale with. However, the error maps show that LAStools and ENVI exhibit large error values compared to ours. This is most apparent at the top right corner and at the bottom of the scene. We believe this is due to surface interpolation around misclassified ground points. On the other hand, the errors from our method exhibit lower error values due to the localized predictions. This observation is supported by the standard deviation values shown in Table~\ref{tab:results}. Our method achieved the lowest standard deviation of 1.1m, while LAStools and ENVI achieved 1.23m and 1.4m respectively. 
It is interesting to note that our network was able to achieve low error values for points representing the roadways (see error map) without having to initial classify them. This indicates that the network has the ability to learn that road regions are good candidates for bare-Earth points.

\begin{figure*}[tp]
\begin{center}
\includegraphics[angle=90,width=13cm]{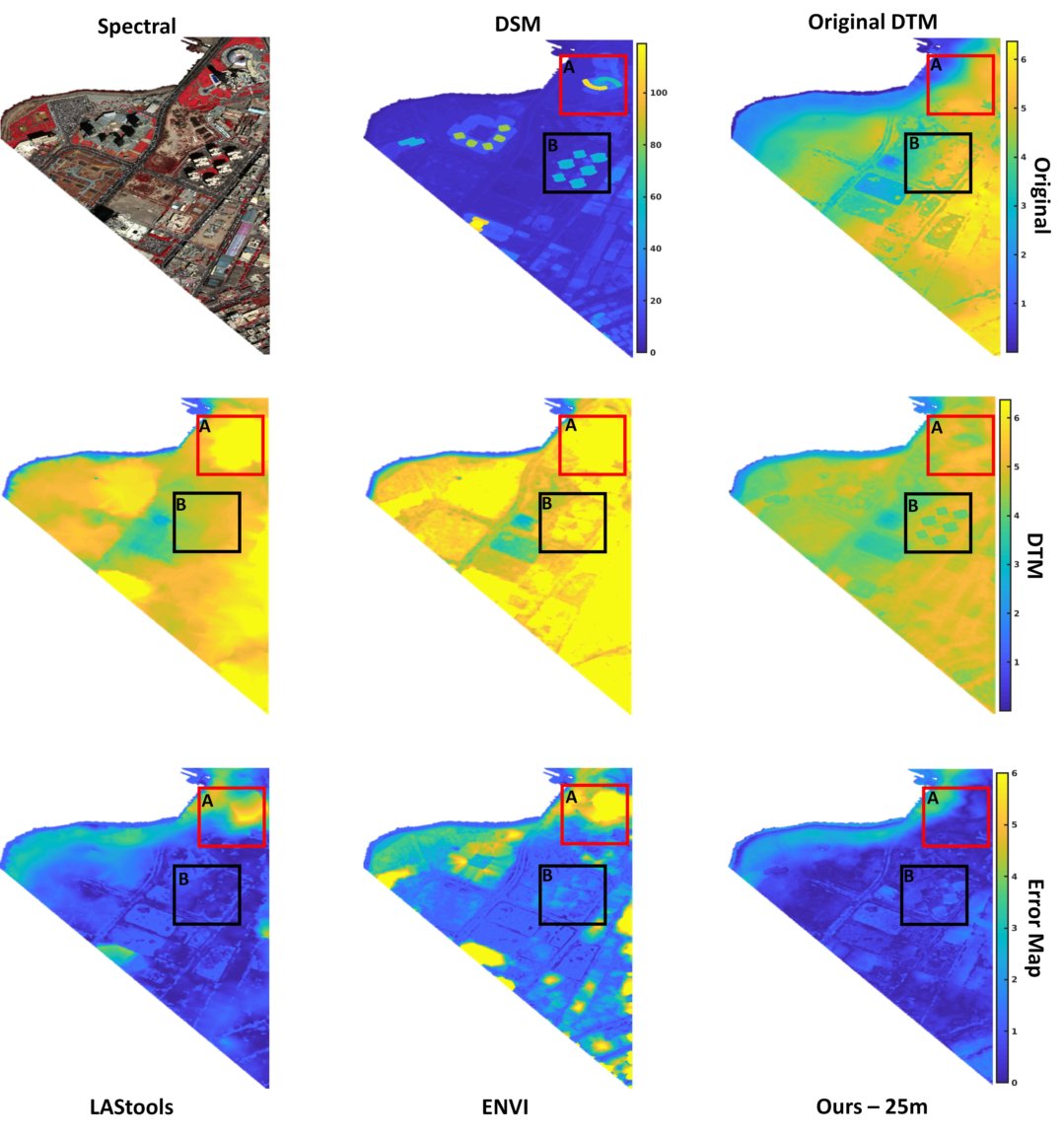}
\end{center}
\caption{Comparison of results for the Tripoli high-rise scene. The first row, from left to right, shows the original color point cloud and the corresponding DSM and ground truth DTM. The second row shows the estimated DTM obtained using LAStools, ENVI, and our method. By subtracting these from the ground truth (top-right) we obtain the corresponding error maps for each method (bottom row). Red and black boxes are marked with `A` and `B` respectively for printing purposes. }
\label{fig:Highrise_Results}
\end{figure*}

 Finally, for the high-rise scene in Figure \ref{fig:Highrise_Results}, we scored the lowest MAE of 0.77m, outperforming LAStools and ENVI with MAE values of 1.22m and 2.42m, respectively. The predicted DTM maps for both LAStools and ENVI show large areas of saturated values (\ie~values larger than the highest ground truth value) compared to our predictions. The error maps show that ENVI struggled most at the edges compared to LAStools and our predictions, with ours showing the lowest error. The error maps also show that both LAStools and ENVI exhibit high error values around unconventional structures (see arc-like buildings within the red box), while our method was able to achieve a low error for the same region. The middle region (marked by the black box) shows that our DTM predictions at the locations of the high-rise buildings are lower than the ground truth. However, by examining the ground truth DTM, one can notice that our estimated elevation for the building footprints actually matches the surrounding areas within the ground truth DTM.  
 

 In summary, the results show that our method can achieve an overall average prediction error of 11.5\% (\ie~dividing the MAE by the maximum value of the ground truth DTM) without post-processing, while LAStools and ENVI achieved an error of 16\% and 29\%, respectively.
 
 \subsection{Effect of Neighborhood Size}
\label{sec:effect_of_neighborhood_size}

In Section \ref{subsection:Neighborhood} we described the importance of choosing a neighborhood radius $R$, in order to learn contextual information. To evaluate the effect of neighborhood radius on DTM results, we trained four different networks, each with a different neighborhood size: 5m, 12m, 18m, and 25m. Larger radii were not included due to memory limitation. The quantitative results for this experiment were reported without any post-processing as well, to allow for a fair comparison and eliminate factors that might bias the results.
The quantitative results are shown in Table \ref{table:scale_exp}.

\begin{table}[!htbp]
\centering
\caption{The effect of varying the neighborhood search radii on the corresponding MAE and $\sigma$ values using the ISPRS data.}
\label{table:scale_exp}
\begin{tabular}{c|c|c}
\textbf{Radius} & \textbf{MAE} & \textbf{$\sigma$}  \\ \hline 
5m            & 3.0m                   & 2.10m \\        
12m           & 1.2m                   & 0.92m \\        
18m           & 1.0m                   & 0.71m \\        
25m           & 0.7m                   & 0.75m \\ \hline 
\end{tabular}
\end{table}

\begin{figure*}[tp]
\begin{center}
\includegraphics[angle=90,width=13cm]{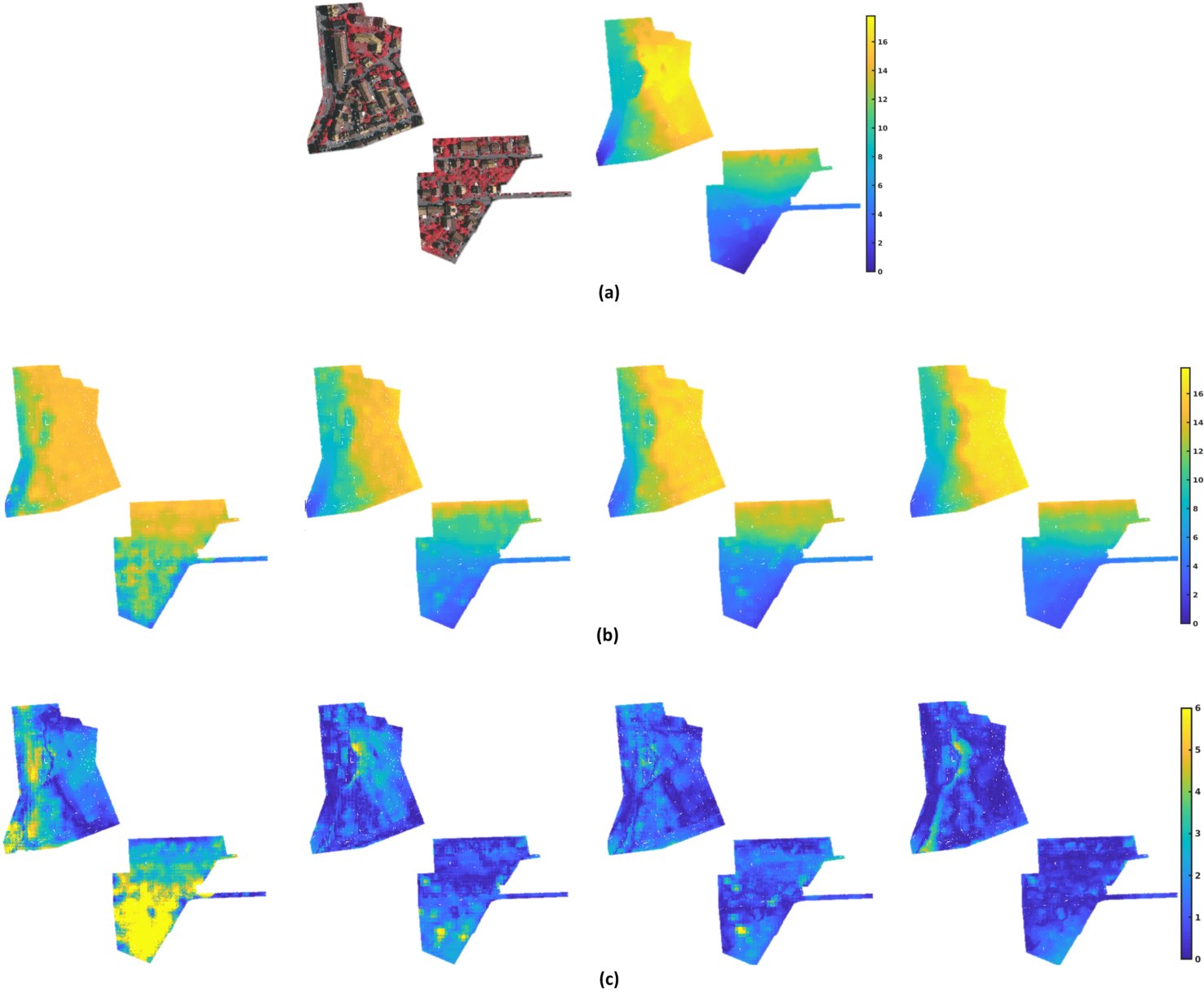}
\end{center}
\caption{In (a), the figure shows the original colored point cloud and the corresponding DTM map. In (b), the predicted DTMs are shown for 5m, 12m, 18m, and 25m radii. In (c), it shows the corresponding error maps.}
\label{fig:scale_exp}
\end{figure*}

 Figure~\ref{fig:scale_exp}, shows the predicted DTM for the different neighborhoods radii, and the corresponding error map.
 The result from the 5m radius contains the most noise among the different neighborhood sizes. However, increasing the search radius to 12m and 18m shows better predictions with much cleaner results compared to the 5m neighborhood. Finally, the result for the 25m radius shows the best overall prediction among all. While the 5m setting showed the worst results, it very important as it helps us interpret the network's behavior. 
 
 Table~\ref{table:scale_exp} shows the MAE and the $\sigma$ for each neighborhood size. By examining Figure~\ref{fig:scale_exp}, the error map for the 5m radius shows the highest overall error values in many regions with a MAE of 3m.  However, such regions are mostly buildings (see Figure \ref{fig:results-overview} for reference DSM and colorized point cloud). This is interesting as it indicates that having a small radius is not sufficient to capture the changes in the terrain. Instead, with such a small radius, the neighborhood points will only describe the surface they cover. Increasing the radius allows larger neighborhoods to potentially include terrain points, as shown for the 12m and 18m radii, with MAE of 1.2m, and 1m respectively. Using a 25m radius allow the neighborhood region to contain a larger area than the 12m and 18m radii, therefore, resulting in lower error values with a MAE of 0.7m.
 

\section{Conclusions}
\label{sec:conclusions}

In the is paper, we introduced a method to estimate the DTM, bypassing the classification stage and operating directly on the point cloud. Our method extends previous work by learning neighborhood information and seamlessly integrating this with point-wise and block-wise global features. 
We achieved a lower overall error compared two widely used software packages for DTM extraction, namely LAStools and ENVI. 
Furthermore, we showed that using a larger neighborhood area results in a lower error as compared to a smaller neighborhood. 
Future work will address the following remaining gaps: 
First, while our results show promising performance, it is important to collect more data in order to thoroughly test this approach. Furthermore, incorporating more data into the training process, will enable our network to handle more diverse scenes. 
Our vision is to develop a unified framework that can use a single network to infer the DTM for  solutions at scale. 
This will benefit both the average user and geoprocessing pipelines by eliminating the need 
to manually adjust scene-specific parameters. Second, in this paper, our network was trained using a direct mapping to a ground truth DTM. We are currently investigating another loss function that may improve our DTM estimation by introducing a correction factor using other point cloud products (\eg~nDSM). Finally, future work should study the effect of combining other tasks (such as segmentation) with our DTM estimation framework, as well as investigating the use of unsupervised learning for terrain modeling.





\bibliographystyle{elsarticle-harv} 
\bibliography{main}





\end{document}